\documentclass[lettersize,journal]{IEEEtran}
\usepackage{cite}
\usepackage{hyperref}       
\usepackage{url}            
\usepackage{booktabs}       
\usepackage{amsfonts}       
\usepackage{nicefrac}       
\usepackage{microtype}      
\usepackage[table]{xcolor}         
\usepackage{amsmath, amssymb, amsthm}
\usepackage[ruled, vlined]{algorithm2e}
\usepackage{graphicx}
\usepackage{subcaption}
\usepackage{multirow}
\usepackage{threeparttable}
\usepackage{marvosym}
\usepackage{pifont}

\newcommand{\CROSS}{\textcolor{red}{\ding{55}}}
\newcommand{\CHECK}{\textcolor{green}{\checkmark}}

\definecolor{lightgreen}{RGB}{210,255,210}
\definecolor{lightred}{RGB}{255,210,210}  
\definecolor{lightyellow}{RGB}{255,255,210}

\title{GMatch: A Lightweight, Geometry-Constrained Keypoint Matcher for Zero-Shot 6DoF Pose Estimation in Robotic Grasp Tasks}

\author{
    Ming Yang$^{1,2}$,
    and Haoran Li\textsuperscript{1,2,\Letter},
    \thanks{$^{1}$The State Key Laboratory of Multimodal Artificial Intelligence Systems, Institute of Automation, Chinese Academy of Sciences, Beijing 100190, China.}
    \thanks{$^{2}$School of Artificial Intelligence, University of Chinese Academy of Sciences, Beijing, China.}
    \thanks{\textsuperscript{\Letter}Corresponding to lihaoran2015@ia.ac.cn}
}

\newtheorem{lemma}{Lemma}
\newtheorem{proposition}{Proposition}

\newcommand{\tb}[1]{\tilde{\mathbf{#1}}}

\begin{document}

\maketitle

\begin{abstract}
  6DoF object pose estimation is fundamental to robotic grasp tasks. While recent learning-based methods achieve high accuracy, their computational demands hinder deployment on resource-constrained mobile platforms. 
  In this work, we revisit the classical keypoint matching paradigm and propose GMatch, a lightweight, geometry-constrained keypoint matcher that can run efficiently on embedded CPU-only platforms. 
  GMatch works with keypoint descriptors and it uses a set of geometric constraints to establishes inherent ambiguities between features extracted by descriptors, thus giving a globally consistent correspondences from which 6DoF pose can be easily solved. 
  We benchmark GMatch on the HOPE and YCB-Video datasets, where our method beats existing keypoint matchers (both feature-based and geometry-based) among three commonly used descriptors and approaches the SOTA zero-shot method on texture-rich objects with much more humble devices. The method is further deployed on a LoCoBot mobile manipulator, enabling a one-shot grasp pipeline that demonstrates high task success rates in real-world experiments. 
  In a word, by its lightweight and white-box nature, GMatch offers a practical solution for resource-limited robotic systems, and although currently bottlenecked by descriptor quality, the framework presents a promising direction towards robust yet efficient pose estimation. 

  Code will be released soon under Mozilla Public License.
\end{abstract}


\section{Introduction}

Robotic grasp has many useful application scenarios like industrial assembly, logistics automation and home service.
To perform a successful grasp, it's fundamental to how far is the target object (translation) and which direction the gripper should approach it from (orientation), which is known as six degrees of freedom (DoF) object pose estimation.
Over the past decade, research focused on learning-based methods. Early works like~\cite{xiang2018posecnn, peng2019pvnet, hai2023scflow} are called \textit{instance-level} for failing to estimate poses of unseen objects in training set. Then comes \textit{category-level} methods like~\cite{wang2019nocs, tian2020nocs, zhou2023nocs, chen2024nocs}, which can generalize within the predefined category, and \textit{zero-shot} methods like~\cite{liu2022gen6d, he2022onepose++, castro2023posematcher, wen2024foundationpose}, which can work for any object that is assiged on run-time. However, with the advances in generalization ability, deeper network~\cite{liu2022gen6d,labbe2022megapose, he2022onepose++}, on-spot rendering~\cite{wen2024foundationpose, nguyen2024gigapose}, and iterative refinement~\cite{castro2023posematcher,cai2024gspose}
are heavily used in current zero-shot methods, which make them hard (if not impossible) to deploy on mobile robots or embedded system, thus hindering their application in robotic grasp tasks.

\begin{figure*}
  \centering
  \includegraphics[width=\textwidth]{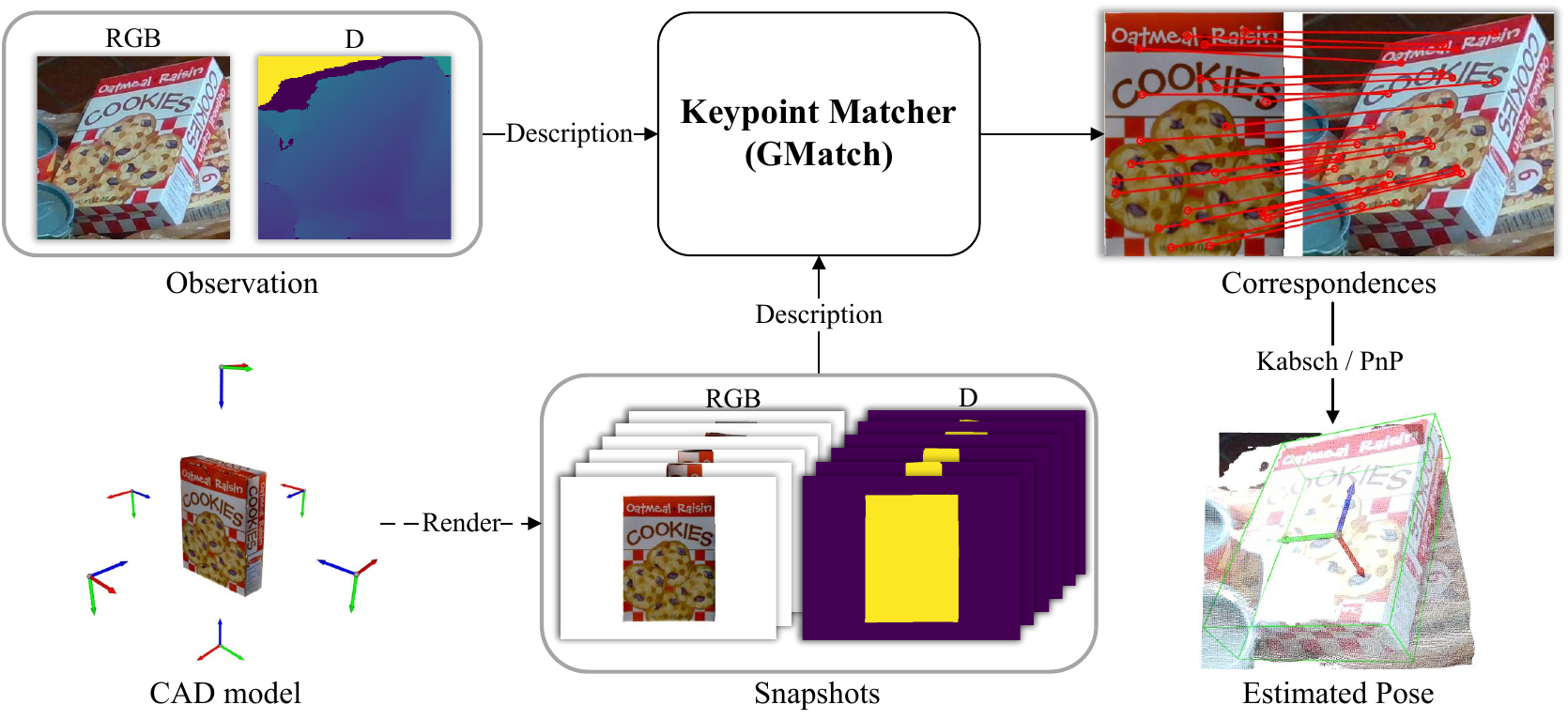}
  \caption{Overview of the matching-based pose estimation pipeline. Given a set of RGB-D images (snapshots) rendered from target CAD model as the source and a scene image (observation) as the target, the descriptor processes them independently to generate keypoints and feature vectors, which are used to reason correspondences by keypoint matcher. Afterwards, Kabsch algorithm~\cite{kabsch1978discussion} or PnP~\cite{gao2003p3p} is used to solve the pose from 3D-3D or 2D-3D correspondences.}
  \label{fig:general_pipeline}
\end{figure*}

In this paper, we revisit the classic pose estimation paradigm based on keypoint matching (see Fig.~\ref{fig:general_pipeline}), whose zero-shot property is easily preserved by object-agnostic keypoint descriptor\footnote{That's because mainstream keypoint descriptor detects and describes keypoints based on neighbour pixels}.
Moreover, The sparsity of keypoints enables efficient keypoint matching and modular framework allows substitution of different keypoint descriptor to balance between performance and efficiency under various scenarios, which makes the pipeline even more suitable for application in real robots.

However, keypoint matchers face a fundamental challenge: \textit{local ambiguity}, which means the keypoint descriptor generates similar feature vectors for different keypoints, usually due to repetitive textures, symmetries, or limited visual diversity.

For conventional matchers that rely solely on feature vectors (e.g., K-Nearest Neighbor with Lowe's ratio test~\cite{lowe2004sift}, mutual filter) or 2D keypoints topology (\cite{sarlin2020superglue, lindenberger2023lightglue}), the 3D structure of keypoints are missing or distorted, thus falling short in extracting \textit{geometrically consistent correspondences}\footnote{meaning the two corresponding point sets can be aligned for every point under a rigid transformation.}. On the other hand, while point cloud registration methods like~\cite{fischler1981ransac} can hold geometric consistency, they have various problems when transferring directly to pose estimation pipeline, e.g., non-deterministic as in RANSAC sampling~\cite{fischler1981ransac}, inefficiency as in MAC full-search~\cite{zhang2023mac}, etc. Moreover, when we need to add some constraints for practical reasons like imaging angle, they fall in short.

In this work, we introduce \textbf{GMatch}, a geometry-aware matching algorithm that reformulates the correspondence problem as an incremental search. It uses a set of geometric characteristics that is provably complete to eliminate local ambiguity and is also open to new constraints. Combined with SIFT descriptor, it performs pose estimation using RGB-D images.
As shown in Table~\ref{tab:hope-results} and Table~\ref{tab:ycbv-results}, our method outperforms all naive combinations of popular feature-based matchers and point cloud registration methods against various descriptors while approches SOTA zero-shot method~\cite{wen2024foundationpose} on texture-rich objects. Deployed in the real robot, it provides pose estimation stable enough for high success rate grasping, thus proving its practical value.

The merits of our algorithm can be summarized as following:
\begin{itemize}
  \item \textbf{Deployability}: No need of professional modeling devices or softwares to get CAD model. Our method itself can act as a quick scanner with RGB-D camera by its per-image matching nature, which is handy to deal with new objects.

  \item \textbf{Simplicity}: It's both white-box and deterministic, and has only one key parameter to tune when adapting to new descriptor (i.e., the feature similarity threshold $\epsilon_f$ in Algorithm~\ref{alg:gmatch-step}).

  \item \textbf{Lightweight and flexibility}: It can run on embeded systems that has no GPU while also adaptive to heavy but powerful descriptors on performant devices.
\end{itemize}

\section{Related Works}

\textbf{Keypoint Descriptors.} Keypoint descriptors are usd to detect sparse and repeatable keypoints from images, and describe them with feature vectors. Generally speaking, handcrafted descriptors (\cite{lowe2004sift,rublee2011orb,calonder2010brief,leutenegger2011brisk,alahi2012freak}) focus on the efficient encoding of local image patches, while learning-based descriptors (\cite{yi2016lift,detone2018superpoint,revaud2019r2d2,tyszkiewicz2020disk,parihar2021rord,zhao2022alike,wang2023awdesc,wang2023featurebooster,xue2023sfd2}) emphasize robustness under challenging conditions, such as poor lighting, blur, and occlusion. In our case, SIFT balances performance and efficiency the best, and we note the resulting pose estimation algorithm as GMatch-SIFT.

\textbf{Keypoint Matchers.} In RGB-D perception, 2D keypoints extracted by descriptors can be reconstructed to 3D keypoints with depth, which provides keypoint matchers more information to generate correspondences. Despite simplicity, nearest neighbour combined with Lowe's ratio test~\cite{lowe2004sift} use only feature vectors. And 2D keypoints that recent learning-based methods (\cite{sarlin2020superglue,lindenberger2023lightglue,sun2021loftr,chen2022aspanformer,zhu2023pmatch}) use with features are actually distorted in their relative position due to imaging. On the other hand, while point cloud registration methods (\cite{fischler1981ransac,chum2003loransac,zhang2023mac,yang2021teaser,huang2021predator}) use 3D keypoints and are good at preserving intrinsic geometries, their full-search stategy are time-consuming (usually quadratic or even exponential to candidate correspondences size).
GMatch addresses these limitations by generating hypotheses with feature vectors and checking geometric characteristics of 3D keypoints, which eliminates local ambiguities in candidate correspondences with linear time complexity.

\textbf{Pose Estimation in Robotic Grasp.}
Robotic manipulation has been one of the most important downstream tasks of pose estimation and many researches are done to serve the purpose.
DOPE~\cite{tremblay2018dope} and Sim2Real Pose~\cite{chen2022sim2realpose} are trained on synthetic data generated by render engines first and transferred to reality by domain adaptation or randomization, leading to a instance-level robotic manipulation. DGPF6D~\cite{liu2024dgpf6d} uses contrastive learning framework to achieve category-level pose estimation and performs picking on various objects with a Yaskawa robotarm. FoundationPose~\cite{wen2024foundationpose} and MegaPose~\cite{labbe2022megapose} also demonstrate zero-shot methods potential by their highly accurate pose tracking and grasping. However, we observe that their hardware, ranging from NVIDIA TITAN X to RTX3090, are luxurious for embedded systems, thereby only suitable for desktop manipulation with fixed roboarm.

\section{Methodology}

Given two point sets and related feature vectors extracted from the source and target images, the objective of GMatch is to find geometrically consistent correspondences within candidate pairs given by feature similarity judgment. In this section, we first propose two choices of numerical characteristics in Sec.~\ref{subsec:key-idea}, and then give the searching-based procedure to enforce these geometric constraints in Sec.~\ref{subsec:method}.

\begin{figure*}
  \centering
  \includegraphics[width=\textwidth]{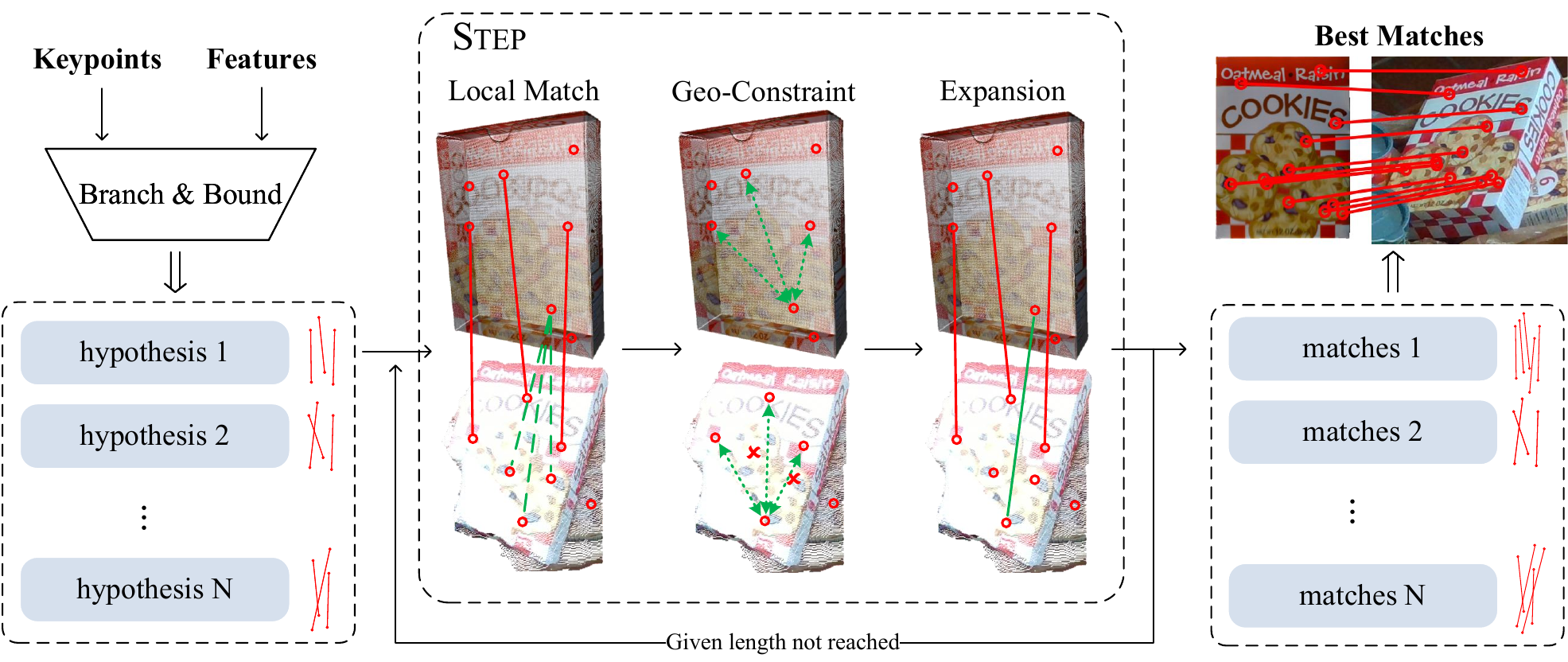}
  \caption{GMatch performs incremental search (\textsc{Step}) over hypothese generated by branch-and-bound stategy and select the matches with the max length as output. In the illustrated example with repetitive grape textures, three locally plausible candidate pairs are initially identified. GMatch filters out inconsistent pairs using geometric characteristics such as relative distance and scalar triple product, retaining only globally consistent correspondences.}
  \label{fig:gmatch_pipeline}
\end{figure*}

\subsection{Key Idea} \label{subsec:key-idea}

As an incremental search algorithm, the key idea of GMatch is to define numerical characteristics of a point set and check them at each step of adding candidate pairs. Therefore, it's flexible to add or replace constraints when new assumptions are adopted as presented below.

We first show our most important characteristics, pairwise distance, which assure us an orthogonal matrix and translation vector.

\begin{lemma} [Satorras et al.~\cite{satorras2021proof}] \label{lemma:pairwise-distance}
  Given two ordered point sets $\{\mathbf{x}_i\}_{i=1}^n, \{\mathbf{y}_i\}_{i=1}^n \subset \mathbb{R}^3$ satisfying $ \lVert \mathbf{x}_i - \mathbf{x}_j \rVert = \lVert \mathbf{y}_i - \mathbf{y}_j \rVert, \quad \forall i,j = 1,\dots,n$, there exists an orthogonal matrix $\mathbf{Q} \in \mathbb{R}^{3 \times 3}$ and a translation vector $\mathbf{t} \in \mathbb{R}^3$ such that $\mathbf{y}_i = \mathbf{Q}\mathbf{x}_i + \mathbf{t}$ for all $i$.
\end{lemma}
\begin{proof}
  see Appendix~\ref{appendix:proofs}.
\end{proof}

Since $\mathbf{Q}$ can have determinant $\pm 1$, this formulation alone cannot distinguish between rotation (det$(\mathbf{Q}) = +1$) and reflection (det$(\mathbf{Q}) = -1$), which is also know as \text{chirality issue}. To address this, our first choice is to use scalar triple product as complementary.

\begin{proposition} \label{prop:equivalence}
  For any $n > 0$, two ordered point sets
  $\{\mathbf{x}_i\}_{i=1}^n, \{\mathbf{y}_i\}_{i=1}^n \subset \mathbb{R}^3$
  are geometrically consistent.
  $\iff \forall\, i,j,k,\ell$,
  \[
    \lVert x_i - x_j \rVert = \lVert y_i - y_j \rVert,
  \]
  and
  \[
    (x_i - x_j) \times (x_i - x_k) \cdot (x_i - x_\ell)
    =
    (y_i - y_j) \times (y_i - y_k) \cdot (y_i - y_\ell)
  \]
\end{proposition}
\begin{proof}
  see Appendix~\ref{appendix:proofs}.
\end{proof}

\begin{figure}
  \centering
  \includegraphics[width=0.85\linewidth]{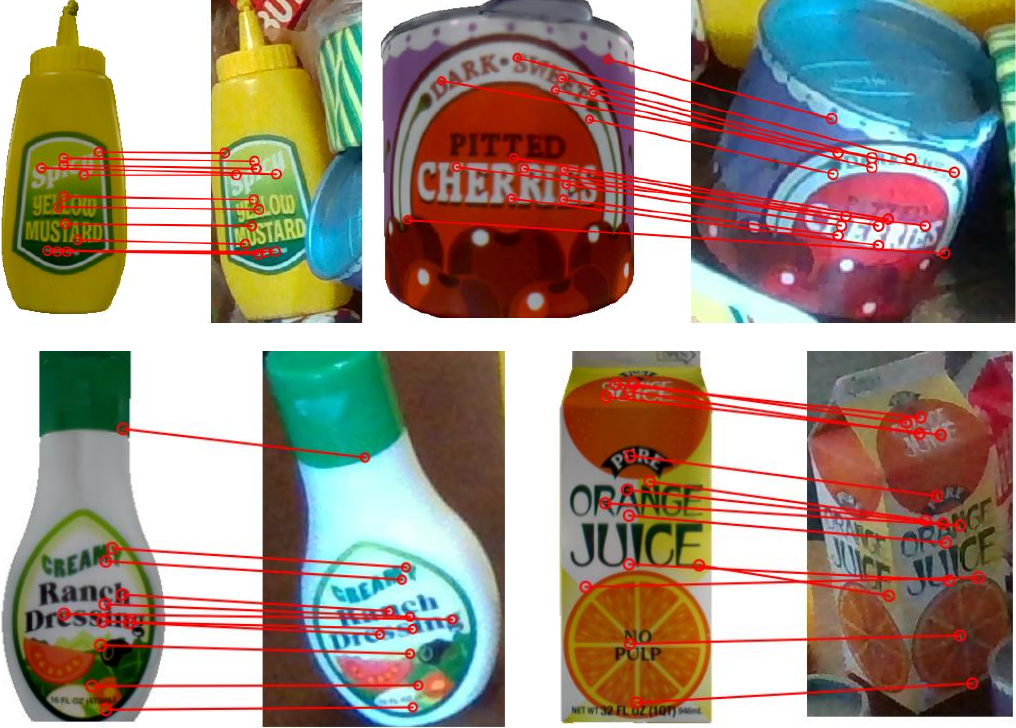}
  \caption{Dense keypoints with alike features are extracted on flat or approximately flat text region, which yields many plausible matches that leads to flip-over.}
  \label{fig:flip-over demo}
\end{figure}

Proposition~\ref{prop:equivalence} reveals the theoretical value of pairwise distance and triple product. By checking these two characteristics, we are guaranteed the geometric consistency of resulting correspondences. Despite theoretical completeness, this choice doesn't always work as expected in real-world settings due to following reasons. The obvious one is, enumeration of four pairs in correspondences is too expensive for embedded devices. \footnote{Unless you use triple product's equivalent form, off-plane distance, which maintains a plane since search starts and check the distance between the plane and the first off-plane pair.} Besides, given opaqueness of target objects, we may want to constrain the result not to \textit{filp over}, which can be quite common when objectes have flat surface (see Fig.~\ref{fig:flip-over demo}).

To resolve this practical issue, we introduce an opacity assumption. Specifically, we assume that any triangle formed by three keypoints is only visible from a single side. Formally, given three points $\{ \mathrm{pt}_i \}_{i=1}^3 \subset \mathbb{R}^3$ and a camera viewing direction $\mathrm{view}$, we require the following term
\[
  \mathrm{sign}\left( \left( \mathrm{pt}_1 - \mathrm{pt}_2 \right) \times \left( \mathrm{pt}_1 - \mathrm{pt}_3 \right) \cdot \mathrm{view} \right)
\]
to remain consistent.\footnote{Note that we only need to check any three non-colinear keypoints instead of all combinations.}

In general, we demonstrate two choices of numerical characteristics. The first one is pairwise distance and triple product, which has excellent theoretical property. The other is based on practical concerns and performs better in reality.

\subsection{Method} \label{subsec:method}

Briefly, GMatch chooses top-$T$ similar pairs from candidates and use branch-and-bound to genrate hypothese of length 3 (the minimal length of 3D-3D correspondences to determine transformation except the colinear case). Note that constraints are used in branch-and-bound stage to ensure geometric consistency of hypothese, unlike random strategy of RANSAC. Then, GMatch expand each hypothesis one pair per step, where the one with minimal cost are chosen from pairs satisfying constraints. See Fig.~\ref{fig:gmatch_pipeline} for illustration.

Before jumping into details, we first do some notation and definition work. Let $\mathrm{cld}^s$ and $\mathrm{cld}^t$ denote the point clouds reconstructed from the source and target depth images. Assume that a keypoint descriptor extracts $n^s$ and $n^t$ keypoints from $\mathrm{img}^s$ and $\mathrm{img}^t$, respectively. Their pixel coordinates are denoted by
$\{\mathrm{pix}^s_i\}_{i=1}^{n^s}, \{\mathrm{pix}^t_i\}_{i=1}^{n^t} \subset \mathbb{Z}^2$,
and the associated feature vectors by
$\{\mathrm{feat}^s_i\}_{i=1}^{n^s}, \{\mathrm{feat}^t_i\}_{i=1}^{n^t}$.
By indexing into the point clouds using the pixel locations, we obtain their 3D coordinates
$\{\mathrm{pt}^s_i\}_{i=1}^{n^s}, \{\mathrm{pt}^t_i\}_{i=1}^{n^t} \subset \mathbb{R}^3$.

The distance of feature vectors is denoted as $d_f(\cdot, \cdot)$, whose choice depends on the specific descriptor.
The distance matrix cost function $g$ quantifies the inconsistency introduced when adding a candidate correspondence to the current match set. It is defined as the maximum pairwise error with respect to all existing matches:
\[
  g(\mathrm{matches}, \mathrm{pair})
  = \max_{p \in \mathrm{matches}} \; \delta(p, \mathrm{pair}),
\]
where the pairwise error term $\delta$ is the relative error ratio \footnote{This term penalizes candidate pairs formed by points that are too close to each other, since such pairs contribute little to improving pose estimation accuracy.} with hard margin $\eta$ \footnote{The tolerance $\eta$ accounts for depth sensor noise, which may introduce small deviations in measured distances.}.

\[
  \delta(p, \mathrm{pair}) =
  \begin{cases}
    \dfrac{\lvert l^s - l^t \rvert}{l^s}, & \text{if } \lvert l^s - l^t \rvert < \eta, \\[1.2ex]
    1,                                    & \text{otherwise}.
  \end{cases}
\]
Here,
$
  l^s = \big\lVert \mathrm{pt}^s_{i_1} - \mathrm{pt}^s_{i_2} \big\rVert,
  l^t = \big\lVert \mathrm{pt}^t_{j_1} - \mathrm{pt}^t_{j_2} \big\rVert,
$
with $p = (i_1, j_1)$ and $\mathrm{pair} = (i_2, j_2)$.

Using these symbols ($\mathrm{feat}, \mathrm{pt}, d_f, g$), we present \textsc{Step} of GMatch as Algorithm~\ref{alg:gmatch-step}.

\begin{algorithm}
  \caption{GMatch-\textsc{Step}} \label{alg:gmatch-step}
  \KwIn{
    A list of currently matched pairs $\mathrm{matches}$;
    Feature similarity threshold $\epsilon_f$;
    Geometric cost tolerance $\epsilon_c$;
    View directions $\mathrm{view}^s, \mathrm{view}^t$ of source and target cameras w.r.t their respective coordinate systems, typical $[0, 0, 1]$ since the z-axis aligns with the view direction.
  }
  \KwOut{New match $m$ if found; otherwise $\mathrm{None}$.}

  $\mathrm{candidates} \gets \left\{ (i,j) \, \middle| \, d_f(\mathrm{feat}^s_i, \mathrm{feat}^t_j) < \epsilon_f \right\}$\;

  \tcp{Apply geometric constraint 1: distance matrix cost}
  \For{$\mathrm{pair}$ \textbf{in} $\mathrm{candidates}$}{
    \If{$g(\mathrm{matches}, \mathrm{pair}) > \epsilon_c$}{
      Remove $\mathrm{pair}$ from $\mathrm{candidates}$\;
    }
  }

  \tcp{Apply geometric constraint 2: flip-over removal}
  $(i_1, j_1) \gets \mathrm{matches}[-1]$ \tcp*{last element}
  $(i_2, j_2) \gets \mathrm{matches}[-2]$ \tcp*{second-last element}

  \For{$\mathrm{pair}$ \textbf{in} $\mathrm{candidates}$}{
    $(i_3, j_3) \gets \mathrm{pair}$\;
    $\mathrm{norm}^s \gets (\mathrm{pt}^s_{i_1} - \mathrm{pt}^s_{i_2}) \times (\mathrm{pt}^s_{i_1} - \mathrm{pt}^s_{i_3})$\;
    $\mathrm{norm}^t \gets (\mathrm{pt}^t_{j_1} - \mathrm{pt}^t_{j_2}) \times (\mathrm{pt}^t_{j_1} - \mathrm{pt}^t_{j_3})$\;
    \If{$\mathrm{sign}(\mathrm{norm}^s \cdot \mathrm{view}^s) \neq \mathrm{sign}(\mathrm{norm}^t \cdot \mathrm{view}^t)$}{
      Remove $\mathrm{pair}$ from $\mathrm{candidates}$\;
    }
  }

  \Return{$\underset{p \in \mathrm{candidates}}{\arg\min}\ g(\mathrm{matches}, p)$ if $\mathrm{candidates} \neq \varnothing$; otherwise $\mathrm{None}$.}
\end{algorithm}

\section{Experiments}

\subsection{Dataset and Setup}
We consider two datasets: YCB-Video and HOPE. YCB-Video consists of household objects that differ in texture richness, and multiple scenes that have different levels of occlusion. On the contrary, HOPE consists of texture-rich objects but offers offers cluttered scens with challenging lighting setttings, including backlighting and angled lighting with cast shadows.
With overall tests covering texture richness, occlusion and lighting condition, we want to justify that GMatch indeed fits into pose estimation better compared with previous keypoint matchers, and help readers understand in what cases our method may fail and why it would.

These two datasets are publicly avaliable on BOP platform~\cite{hodan2020bop}, and we use its evaluation toolkit and online judge to make our results repeatable and convincing. Following our baseline protocols, we use following metrics:
\begin{itemize}
  \item Area under the curve (AUC) of ADD and ADD-S~\cite{xiang2018posecnn}.
  \item Average recall (AR) of VSD, MSSD and MSPD metrics introduced in the BOP challenge~\cite{hodan2020bop}.
\end{itemize}

The default settings are listed here: We use Euclidean distance as $d_f$ for SIFT and SuperPoint, and Hamming distance for ORB, with their feature threshold $\epsilon_f$ being $0.1$, $1.25$ and $90$. ICP~\cite{chen1991icp} is used as the downstream refiner for all keypoint matcher. GMatch-specific parameters are that $\epsilon_c=0.08, T=24$ and max search length $L=24$.

\subsection{Results on HOPE}

\paragraph{Baselines} We use feature-based matcher (\cite{lowe2004sift,lindenberger2023lightglue}) and point cloud registration method (\cite{fischler1981ransac,yang2021teaser}) as baselines to compare with GMatch, and provide two learning-based methods (\cite{labbe2020cosypose,tremblay2018dope}) as references. We use the official github repository for SuperPoint, LightGlue and TEASER++~\cite{yang2021teaser}, and OpenCV implementation for SIFT and ORB. We implement RANSAC in Python and release it with our code. CostPose and DOPE results are adopted from BOP platform.

\begin{table}
  \caption{Pose estimation results measured by AR scores (MSPD, MSSD, VSD) on the HOPE dataset (\%). NN denotes Nearest Neighbor matching with Lowe's ratio test (threshold = 0.75); SPP denotes SuperPoint~\cite{detone2018superpoint}.}
  \centering
  \setlength{\tabcolsep}{5pt}
  \begin{threeparttable}
    \begin{tabular}{ll|c|cccc}
      \toprule

      \multicolumn{2}{c|}{Methods}                           & Zero-shot                                  & MSPD             & MSSD          & VSD              & Avg.                                \\

      \midrule

      \multirow{2}{*}{SPP}                                   & LightGlue~\cite{lindenberger2023lightglue} & \CHECK           & 24.9          & 21.6             & 31.9             & 26.1             \\
                                                             & \textbf{Ours}                              & \CHECK           & 34.0          & 28.3             & 49.7             & 37.3             \\

      \midrule

      \multirow{2}{*}{ORB}                                   & NN~\cite{lowe2004sift}                     & \CHECK           & 30.5          & 26.4             & 37.2             & 31.4             \\
                                                             & \textbf{Ours}                              & \CHECK           & 47.9          & 42.5             & 57.1             & 49.1             \\

      \midrule

      \multirow{5}{*}{SIFT}                                  & NN                                         & \CHECK           & 49.7          & 44.8             & 52.6             & 49.0             \\
                                                             & LightGlue                                  & \CHECK           & 52.2          & 47.9             & 56.4             & 52.1             \\
                                                             & RANSAC~\cite{fischler1981ransac}           & \CHECK           & 55.0          & 50.1             & 57.6             & 54.2             \\
                                                             & TEASER++~\cite{yang2021teaser}             & \CHECK           & 58.1          & 52.9             & 59.2             & 56.8             \\
                                                             & \textbf{Ours}                              & \CHECK           & \textbf{64.0} & \underline{57.9} & \underline{67.8} & \underline{63.2} \\

      \midrule

      \multicolumn{2}{c|}{CosyPose~\cite{labbe2020cosypose}} & \CROSS                                     & \underline{62.9} & \textbf{59.4} & \textbf{69.1}    & \textbf{63.8}                       \\
      \multicolumn{2}{c|}{DOPE~\cite{tremblay2018dope}}      & \CROSS                                     & 49.8             & 29.7          & 27.7             & 35.7                                \\

      \bottomrule
    \end{tabular}
  \end{threeparttable}
  \label{tab:hope-results}
\end{table}

\paragraph{Results} Table~\ref{tab:hope-results} presents comparison results. In general, in the  descriptor-matcher pipeline, SIFT performs far better than ORB and SuperPoint, where 15\%--25\% improvements are observed. Therefore, we mainly focus on the SIFT. In that case, point cloud registration methods are slightly better than feature-based ones, while GMatch is better than registration methods (+9.0\% with RANSAC, +6.4\% with TEASER++). Moreover, our method (GMatch-SIFT) achieves CosyPose's accuracy with weaker assumption (zero-shot prediction v.s. instance-level fine-tune). And with the same goal of serving robotic grasp tasks, our method outperforms DOPE significantly (+27.5\%).

\begin{figure}
  \centering
  \includegraphics[width=\columnwidth]{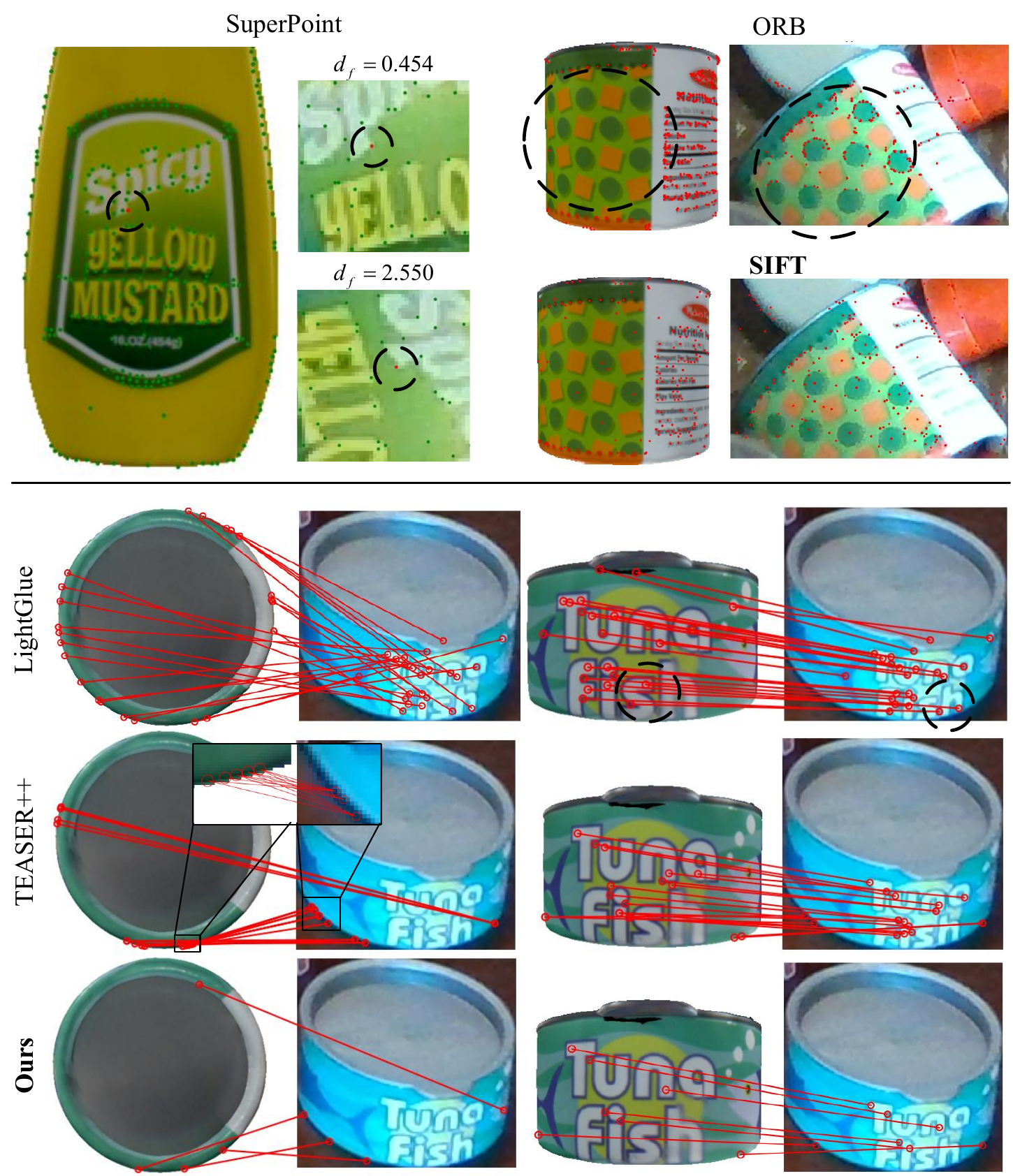}
  \caption{Qualitative comparison: rotation sensitivity for SuperPoint and weak detection repeatability for ORB; inaccurate matches for LightGlue and redundant cross matching for TEASER++.}
  \label{fig:hope-qualitative}
\end{figure}

\paragraph{Qualitative}
Fig.~\ref{fig:hope-qualitative} visualizes different preferences of descriptors and keypoint matchers. Among the three descriptors, SuperPoint is sensitive to in-plane rotation and ORB fails in detecting repeatable keypoints, which makes SIFT a more reasonable choice. For the two representative feature matching and point cloud registration methods, LightGlue fails in preserving underlying 3D structure while TEASER++ is confused by neighbouring cross matching and miss the correct view. Our method are flexible to combine different constraints to preserve geometric consistency and filter out pairs that are close to each other.

\subsection{Results on YCB-Video}

\begin{table}[htbp]
  \setlength{\tabcolsep}{5pt}
  \centering
  \caption{Pose estimation results measured by AUC of ADD/ADD-S on YCB-Video dataset (\%). Background color indicates richness of visible textures (\colorbox{lightgreen}{always}; \colorbox{lightyellow}{sometimes}; \colorbox{lightred}{barely}).}
  \resizebox{\linewidth}{!}{
    \begin{threeparttable}
      \begin{tabular}{l|ccccc}
        \toprule
                                  & PREDATOR                 & LoFTR               & FS6D-DPM          & FoundationPose               & \multirow{2}{*}{\textbf{Ours}} \\
                                  & \cite{huang2021predator} & \cite{sun2021loftr} & \cite{he2022fs6d} & \cite{wen2024foundationpose} &                                \\
        \midrule
        \multirow{2}{*}{GPU-free} & \CROSS                   & \CROSS              & \CROSS            & \CROSS                       & \multirow{2}{*}{\CHECK}        \\
                                  & GTX1080Ti                & RTX2080Ti           & RTX2080Ti         & RTX3090                      &                                \\
        \midrule
        \rowcolor{lightgreen}
        master\_chef\_can*        & 73.0                     & 87.2                & 92.6              & \underline{96.9}             & \textbf{97.4}                  \\
        \rowcolor{lightgreen}
        cracker\_box              & 8.3                      & 25.5                & 24.5              & \textbf{96.2}                & \underline{87.3}               \\
        \rowcolor{lightgreen}
        sugar\_box                & 15.3                     & 13.4                & 43.9              & \underline{87.2}             & \textbf{91.2}                  \\
        \rowcolor{lightyellow}
        tomato\_soup\_can         & 44.4                     & 52.9                & 54.2              & \textbf{93.3}                & \underline{82.4}               \\
        \rowcolor{lightyellow}
        mustard\_bottle           & 5.0                      & 59.0                & \underline{71.1}  & \textbf{97.3}                & 66.7                           \\
        \rowcolor{lightyellow}
        tuna\_fish\_can           & 34.2                     & 55.7                & 53.9              & \textbf{73.7}                & \underline{66.1}               \\
        \rowcolor{lightyellow}
        pudding\_box              & 24.2                     & 68.1                & \underline{79.6}  & \textbf{97.0}                & 68.0                           \\
        \rowcolor{lightgreen}
        gelatin\_box              & 37.5                     & 45.2                & 32.1              & \textbf{97.3}                & \underline{96.4}               \\
        \rowcolor{lightyellow}
        potted\_meat\_can         & 20.9                     & 45.1                & \underline{54.9}  & \textbf{82.3}                & 53.3                           \\
        \rowcolor{lightred}
        banana                    & 9.9                      & 1.6                 & \underline{69.1}  & \textbf{95.4}                & 16.1                           \\
        \rowcolor{lightred}
        pitcher\_base             & 18.1                     & 22.3                & \underline{40.4}  & \textbf{96.6}                & 2.7                            \\
        \rowcolor{lightyellow}
        bleach\_cleanser          & 48.1                     & 16.7                & 44.1              & \textbf{93.3}                & \underline{74.7}               \\
        \rowcolor{lightred}
        bowl*                     & 17.4                     & 1.4                 & 0.9               & \textbf{89.7}                & \underline{76.2}               \\
        \rowcolor{lightred}
        mug                       & 29.5                     & 23.6                & 39.2              & \underline{75.8}             & \textbf{89.6}                  \\
        \rowcolor{lightred}
        power\_drill              & 12.3                     & 1.3                 & 19.8              & \textbf{96.3}                & \underline{36.2}               \\
        \rowcolor{lightred}
        wood\_block*              & 70.5                     & 49.9                & \underline{94.7}  & \textbf{97.4}                & 65.1                           \\
        \rowcolor{lightred}
        scissors                  & 25.0                     & 14.6                & \underline{27.7}  & \textbf{95.5}                & 24.2                           \\
        \rowcolor{lightgreen}
        large\_marker*            & 38.9                     & 8.4                 & 74.2              & \textbf{96.5}                & \underline{93.0}               \\
        \rowcolor{lightred}
        large\_clamp*             & \underline{83.0}         & 24.1                & 82.7              & \textbf{96.9}                & 33.0                           \\
        \rowcolor{lightred}
        extra\_large\_clamp*      & \underline{72.9}         & 15.0                & 65.7              & \textbf{97.6}                & 3.1                            \\
        \rowcolor{lightred}
        foam\_brick*              & 79.2                     & 59.4                & \underline{95.7}  & \textbf{98.1}                & 6.9                            \\
        \bottomrule
      \end{tabular}

      \begin{tablenotes}
        \item[*] denotes symmetric objects that use AUC of ADD-S. All the other objects use AUC of ADD.
      \end{tablenotes}
    \end{threeparttable}
  }
  \label{tab:ycbv-results}
\end{table}

\paragraph{Baselines} we compared our method (GMatch-SIFT) against four different learning-based algorithms on YCB-Video dataset. PREDATOR~\cite{huang2021predator} is a point cloud registration method using deep attention to focus on the overlap region of two point clouds to generate correspondences. LoFTR~\cite{sun2021loftr} is a detector-free feature matcher that generate dense correspondences. FS6D-DPM~\cite{he2022fs6d} uses encoder-decoder architecture to generate features for correspondence reasoning. FoundationPose~\cite{wen2024foundationpose} is the current SOTA zero-shot algorithm with code publicly available.
Baselines results are adopted from \cite{wen2024foundationpose}.

\begin{figure}
  \centering
  \includegraphics[width=\columnwidth]{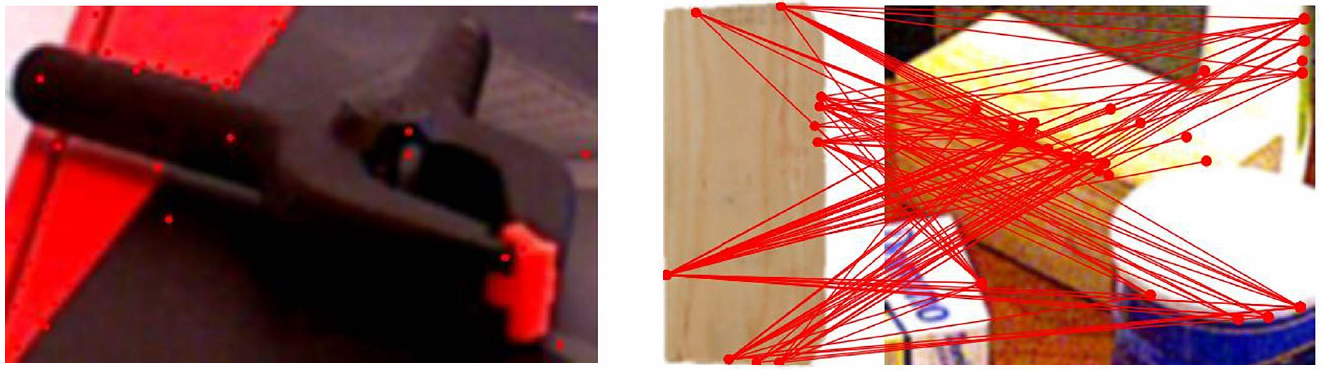}
  \caption{Failure cases of GMatch-SIFT. SIFT detects few keypoints (left) or indistinguishable features (right) on texture-weak objects. The former leaves GMatch no candidate pairs, and the latter often yields plenty of plausible solutions with lower cost than the real one.}
  \label{fig:failure-casees}
\end{figure}

\paragraph{Results} Table~\ref{tab:ycbv-results} shows accuracy comparison per object. On texture-rich objects, SIFT detects keypoints and describe them with nearby texture, followed by GMatch extracting globally consistent matches. In that case (lightgreen rows), our method approaches or achieves SOTA method, and significantly outperforms other three correspondences-based methods. However, on objects that textures can only be seen from certain views (lightyellow), our method's performance becomes unstable and will finally fail on objects without textures (lightred)\footnote{Note that the abnormal high performance on bowl (76.2\%) and mug (80.6\%) owes to ICP refinement instead of our method.}. Fig.~\ref{fig:failure-casees} illustrates how SIFT fails and bottlenecks GMatch.

\paragraph{Runtimes}
Table~\ref{tab:modular-runtimes} shows the modular runtimes of GMatch-SIFT. While the source image description takes up 190~ms, it needs to be done once for a certain object by caching the extracted keypoints and features. Specially, we underline the astonishingly low latency of GMatch considering it runs on consumer-level CPU. It eables GMatch-SIFT to run under sequential inputs with runtimes around 100~ms (w/o ICP) or 300~ms (w/ ICP).

\begin{table}
  \caption{Modular runtimes of GMatch-SIFT (sec). Tested on i5-12400F with 8GB memory. We report the average of the 5 texture-rich objects of YCB-Video dataset and neglect texture-weak objects for they run abnormally faster due to lack of keypoints.}
  \centering
  \setlength{\tabcolsep}{5pt}
  \resizebox{\linewidth}{!}{
    \begin{threeparttable}
      \begin{tabular}{ccccc|c}
        \toprule
        Src. Desc. & Targ. Desc. & Feat. Simi. Comput. & GMatch & ICP   & \textbf{Total} \\
        \midrule
        0.190      & 0.012       & 0.057               & 0.012  & 0.212 & 0.483          \\
        \bottomrule
      \end{tabular}
    \end{threeparttable}
  }
  \label{tab:modular-runtimes}
\end{table}

\subsection{Robotic Grasp Tasks}

To further demonstrate the practical value of our method, we build a one-shot grasp pipeline, where CAD model of target object is unnecessary (\textit{model-free}) and only need demonstration once to perform generalized grasp w.r.t. any initial pose (\textit{one-shot}).

\begin{figure}[h]
  \centering
  \includegraphics[width=\columnwidth]{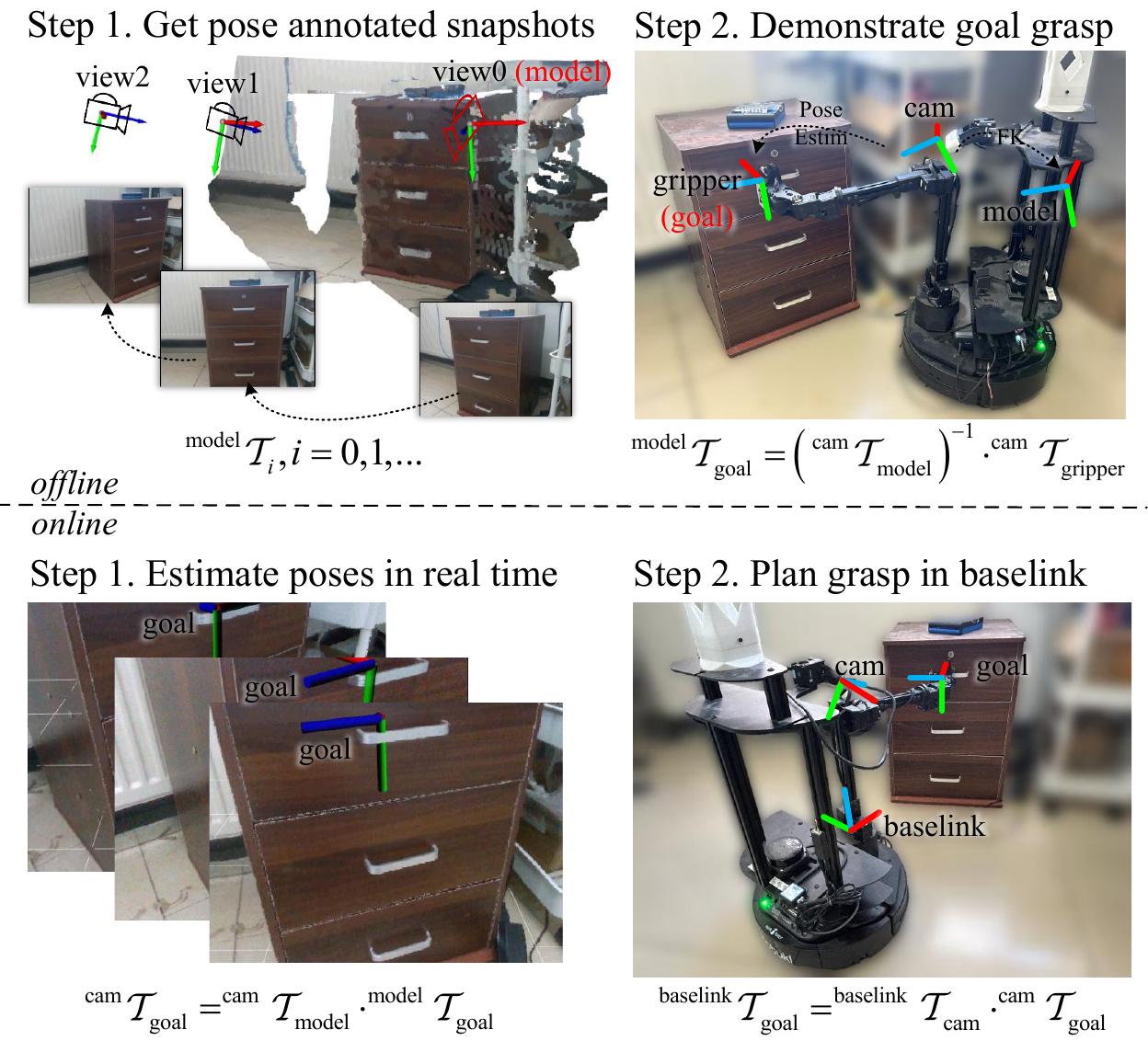}
  \caption{One-shot grasp pipeline with GMatch-SIFT. We get goal grasp on model from one demonstration and can plan grasp from any initial pose afterwards. GMatch-SIFT bridges the gap between the camera coordinate system and the model throughout.}
  \label{fig:grasp-pipeline}
\end{figure}

As illustrated in Fig.~\ref{fig:grasp-pipeline}, our grasp pipeline is comprised of the offline stage and online stage. In the offline stage, we first take RGB-D snapshots around the target object, select a certain view as the model coordinate system and use GMatch-SIFT to annotate poses of the other views. And then we manually align the gripper to the goal grasp pose, from which we can obtain the goal grasp pose w.r.t model coordinate system $^\mathrm{model}\mathcal{T}_\mathrm{goal}$ with estimated pose $^\mathrm{cam}\mathcal{T}_\mathrm{model}$ and eye-hand calibration $^\mathrm{cam}\mathcal{T}_\mathrm{gripper}$. In the online stage, given an arbitrary start position, the robot estimates the goal grasp pose $^\mathrm{cam}\mathcal{T}_\mathrm{goal}$, and then plan the robotic grasp in baselink $^\mathrm{baselink}\mathcal{T}_\mathrm{goal}$.

We test our grasp pipeline on LoCoBot, a 6DoF mobile manipulator equipped with an Intel NUC11 i7-1165G7@2.8GHz and a RealSense D435i. In the offline stage, we take three snapshots for the target object, cabinet and set its handle as the goal grasp. After that, we try ten different initial poses of robot consecutively and based on our observation (as shown in attached video), the robot can always grasp the handle accurately as long as it's reachable for robotic arm.

\section{Limitations and Future Work}
GMatch is essentially a correspondences filter and no new matches or keypoints are added to the initial correspondence set (i.e., the candidate correspondences as mentioned above) during matching. Therefore, it's expected and verified that our method would fail in case that the descriptor fails to extract sufficient keypoints or describe them correctly. Also, missing depth values of keypoints lead to failure in reconstruction to 3D space, which actually makes these keypoints invalid. But this's not worth too much concern because consumer-level depth sensors like Realsense series would suffice based on our experiences.

Since our method are currently bottlenecked by SIFT's incapability to handle occlusion, low-texture and challenging lighting, using superior visual descriptors together with geometric descriptors to exploits would be a promising direction, given the success of FreeZe~\cite{caraffa2024freeze}.

\section{Conclusion}

We propose GMatch, a simple and lightweight matcher that enforces geometric constraints during incremental search, which solves local ambiguity inherent from descriptors. GMatch integrates easily with descriptors such as SIFT to make a zero-shot pose estimation algorithm. Experiments show that our zero-shot pipeline is not only theoretically sound but also performs on par with SOTA learned methods on texture-rich cases. Its practical value is further demonstrated by our one-shot grasp pipeline and high success rate grasping.

\bibliographystyle{IEEEtran}
\bibliography{main}

\appendix

\section{Proofs of Lemmas}
\label{appendix:proofs}
Here we give a more detailed version of the proof by Satorras et al.~\cite{satorras2021proof}, specialized to the $\mathbb{R}^3$ case. We assume all vectors are column vectors; $\mathbf{x}^\top$ denotes transpose, and $\lVert \cdot \rVert$ the Euclidean norm.

\begin{proof} [Proof of Lemma~\ref{lemma:pairwise-distance}]
  Define centered vectors $\tb{x}_i := \mathbf{x}_i - \mathbf{x}_1$ and $\tb{y}_i := \mathbf{y}_i - \mathbf{y}_1$ for $i = 1, \dots, n$. Then:
  \begin{align}
    \lVert \tb{x}_i \rVert            & = \lVert \tb{y}_i \rVert, \label{eq:equal-norm}            \\
    \lVert \tb{x}_i - \tb{x}_j \rVert & = \lVert \tb{y}_i - \tb{y}_j \rVert. \label{eq:equal-dist}
  \end{align}

  Using the identity
  \[
    \tb{x}_i^\top \tb{x}_j = \frac{1}{2} \left( \lVert \tb{x}_i \rVert^2 + \lVert \tb{x}_j \rVert^2 - \lVert \tb{x}_i - \tb{x}_j \rVert^2 \right),
  \]
  and applying \eqref{eq:equal-norm} and \eqref{eq:equal-dist}, we obtain
  \begin{equation}
    \tb{x}_i^\top \tb{x}_j = \tb{y}_i^\top \tb{y}_j. \label{eq:equal-dot}
  \end{equation}

  Therefore, for any $c_1, \dots, c_N \in \mathbb{R}$,
  \begin{align*}
    \left\lVert \sum_i c_i \tb{x}_i \right\rVert^2 & = \left( \sum_i c_i \tb{x}_i \right)^\top \left( \sum_j c_j \tb{x}_j \right)                  \\
                                                   & = \sum_{i,j} c_i c_j \tb{x}_i^\top \tb{x}_j                                                   \\
                                                   & = \sum_{i,j} c_i c_j \tb{y}_i^\top \tb{y}_j = \left\lVert \sum_i c_i \tb{y}_i \right\rVert^2,
  \end{align*}
  which implies
  \begin{equation} \label{eq:equal-express}
    \sum_{i=1}^n c_i \tb{x}_i = 0 \quad \iff \quad \sum_{i=1}^n c_i \tb{y}_i = 0.
  \end{equation}

  Let $\{\tb{x}_{k_1}, \dots, \tb{x}_{k_d}\}$ be a basis of $\mathrm{span}\{\tb{x}_1, \dots, \tb{x}_n\}$ with $d \leq 3$. Then by \eqref{eq:equal-express}, $\{\tb{y}_{k_1}, \dots, \tb{y}_{k_d}\}$ are also linearly independent, and every $\tb{y}_i$ can be expressed as a linear combination of $\{\tb{y}_{k_1}, \dots, \tb{y}_{k_d}\}$ with the same coefficients as for $\tb{x}_i$.

  Apply Gram-Schmidt orthogonalization to $\{\tb{x}_{k_i}\}$ to construct additional vectors $\boldsymbol{\alpha}_1, \dots, \boldsymbol{\alpha}_{3-d}$ such that
  \[
    \boldsymbol{\alpha}_i^\top \boldsymbol{\alpha}_j = \delta_{ij}, \quad \text{and} \quad \boldsymbol{\alpha}_i^\top \tb{x}_{k_j} = 0 \quad \forall i, j,
  \]
  where $\delta_{ij}$ is the Kronecker delta.

  Define $\mathbf{X} \in \mathbb{R}^{3 \times 3}$ as the matrix whose columns are $\tb{x}_{k_1}, \dots, \tb{x}_{k_d}, \boldsymbol{\alpha}_1, \dots, \boldsymbol{\alpha}_{3-d}$, so that
  \begin{equation}
    \mathbf{X}^\top \mathbf{X} =
    \begin{pmatrix}
      (\tb{x}_{k_i}^\top \tb{x}_{k_j})_{d \times d} & \mathbf{0}       \\
      \mathbf{0}                                    & \mathbf{I}_{3-d}
    \end{pmatrix}. \label{eq:XTX}
  \end{equation}
  Similarly, define $\mathbf{Y}$ based on $\tb{y}_i$, satisfying
  \begin{equation}
    \mathbf{Y}^\top \mathbf{Y} =
    \begin{pmatrix}
      (\tb{y}_{k_i}^\top \tb{y}_{k_j})_{d \times d} & \mathbf{0}       \\
      \mathbf{0}                                    & \mathbf{I}_{3-d}
    \end{pmatrix}. \label{eq:YTY}
  \end{equation}

  By \eqref{eq:equal-dot}, \eqref{eq:XTX}, and \eqref{eq:YTY}, it follows that $\mathbf{X}^\top \mathbf{X} = \mathbf{Y}^\top \mathbf{Y}$.
  Since $\mathbf{X}$ and $\mathbf{Y}$ are invertible, define
  \begin{equation} \label{eq:Q-construct}
    \mathbf{Q} := \mathbf{Y}\mathbf{X}^{-1}.
  \end{equation}
  Thus, $\mathbf{Y} = \mathbf{Q}\mathbf{X}$, and particularly,
  \begin{equation} \label{eq:rot-basis}
    \tb{y}_{k_i} = \mathbf{Q} \tb{x}_{k_i}, \quad \forall i = 1, \dots, d.
  \end{equation}
  By linearity and common coefficients, we also have
  \[
    \tb{y}_i = \mathbf{Q} \tb{x}_i, \quad \forall i = 2, \dots, n.
  \]
  Recalling that $\tb{x}_i = \mathbf{x}_i - \mathbf{x}_1$ and $\tb{y}_i = \mathbf{y}_i - \mathbf{y}_1$, we conclude
  \[
    \mathbf{y}_i = \mathbf{Q} \mathbf{x}_i + \mathbf{t}, \quad \text{where} \quad \mathbf{t} := \mathbf{y}_1 - \mathbf{Q}\mathbf{x}_1.
  \]

  It remains to verify that $\mathbf{Q}$ is orthogonal:
  \[
    \mathbf{Q}^\top \mathbf{Q} = (\mathbf{X}^{-1})^\top \mathbf{Y}^\top \mathbf{Y} \mathbf{X}^{-1} = (\mathbf{X}^{-1})^\top \mathbf{X}^\top \mathbf{X} \mathbf{X}^{-1} = \mathbf{I}_n.
  \]
\end{proof}

\begin{proof} [Proof of Proposition~\ref{prop:equivalence}]
  \noindent\textbf{($\Leftarrow$)}
  If $\{\mathbf{x}_i\}_{i=1}^n$ are coplanar, then by the construction in Appendix~\ref{appendix:proofs}, we have
  \[
    d = \dim \operatorname{span}\{\tb{x}_{k_1},\ldots,\tb{x}_{k_d}\} < 3.
  \]
  In this case, if $\mathbf{Y}\mathbf{X}^{-1}$ has determinant $-1$, we can instead use
  \[
    \mathbf{Y}\,\mathrm{diag}(1,1,-1)\,\mathbf{X}^{-1},
  \]
  which still satisfies Eq.~\eqref{eq:rot-basis} and therefore yields a valid rotation matrix in $SO(3)$.

  Otherwise, there exist four points $\mathbf{x}_i,\mathbf{x}_j,\mathbf{x}_k,\mathbf{x}_\ell$ such that
  \[
    V_x = (\mathbf{x}_i - \mathbf{x}_j) \times (\mathbf{x}_i - \mathbf{x}_k) \cdot (\mathbf{x}_i - \mathbf{x}_\ell) \neq 0.
  \]
  By Lemma~\ref{lemma:pairwise-distance}, there exists an orthogonal matrix $\mathbf{Q}$ and translation $\mathbf{t}$ such that
  \[
    \mathbf{y}_i = \mathbf{Q}\mathbf{x}_i + \mathbf{t}.
  \]
  Hence,
  \[
    V_y = (\mathbf{y}_i - \mathbf{y}_j) \times (\mathbf{y}_i - \mathbf{y}_k) \cdot (\mathbf{y}_i - \mathbf{y}_\ell)
    = \det(\mathbf{Q})\,V_x.
  \]
  Since $V_x = V_y$, we conclude that $\det(\mathbf{Q}) = +1$, i.e., $\mathbf{Q} \in SO(3)$.

  \medskip
  \noindent\textbf{($\Rightarrow$)}
  The converse can be verified directly.
\end{proof}

\end{document}